\documentclass[letterpaper, 10 pt, conference]{ieeeconf}  

\IEEEoverridecommandlockouts                              

\newcommand\RL{\mbox{{RL}}}
\newcommand\IFO{\mbox{{IfO}}}
\newcommand\IL{\mbox{{IL}}}
\newcommand\MDP{\mbox{{MDP}}}
\newcommand\LQR{\mbox{{LQR}}}
\newcommand\PILQR{\mbox{{PILQR}}}
\newcommand\DEALIO{\mbox{{DEALIO}}}
\newcommand\PI{\mbox{{PI$^2$}}}
\newcommand\iLQR{\mbox{{iLQR}}}
\newcommand\GAIfO{\mbox{{GAIfO}}}
\newcommand\GAN{\mbox{{GAN}}}
\newcommand\TRPO{\mbox{{TRPO}}}

\newcommand\M{\mathcal{M}}
\renewcommand{\S}{\mathcal{S}}
\newcommand{\A}{\mathcal{A}}
\newcommand\N{\mathcal{N}}

\usepackage{amsmath,amsfonts,amscd} 
\usepackage{algorithm}
\usepackage[noend]{algpseudocode}
\usepackage{amssymb}
\usepackage{mathtools}
\usepackage{tikz}
\usepackage{subfigure}

\usepackage{xcolor}

\title{\LARGE \bf
\DEALIO: Data-Efficient Adversarial Learning for Imitation from Observation}

\author{Faraz Torabi$^{1}$, Garrett Warnell$^{2}$ and Peter Stone$^{3}$
\thanks{*This work has taken place in the Learning Agents Research Group (LARG) at UT Austin. LARG research is supported in part by NSF (CPS-1739964, IIS-1724157, NRI-1925082), ONR (N00014-18-2243), FLI (RFP2-000), ARO (W911NF-19-2-0333), DARPA, Lockheed Martin, GM, and Bosch. Peter Stone serves as the Executive Director of Sony AI America and receives financial compensation for this work. The terms of this arrangement have been reviewed and approved by the University of Texas at Austin in accordance with its policy on objectivity in research.}
\thanks{$^{1}$First Author is with the Computer Science Department, The University of Texas at Austin, USA
        {\tt\small faraztrb@cs.utexas.edu}}%
\thanks{$^{2}$Second Author is with Army Research Laboratory, USA and the Computer Science Department, The University of Texas at Austin, USA
        {\tt\small garrett.a.warnell.civ@mail.mil}}%
    \thanks{$^{2}$Third Author is with the Computer Science Department, The University of Texas at Austin and Sony AI, USA
    {\tt\small pstone@cs.utexas.edu}}%
}

\begin{document}

\maketitle
\thispagestyle{empty}
\pagestyle{empty}

\begin{abstract}
	In imitation learning from observation (\IFO), a learning agent seeks to imitate a demonstrating agent using only {\em observations} of the demonstrated behavior without access to the control signals generated by the demonstrator.
	Recent methods based on adversarial imitation learning have led to state-of-the-art performance on \IFO~problems, but they typically suffer from high sample complexity due to a reliance on data-inefficient, model-free reinforcement learning algorithms.
	This issue makes them impractical to deploy in real-world settings, where gathering samples can incur high costs in terms of time, energy, and risk.
	In this work, we hypothesize that we can incorporate ideas from model-based reinforcement learning with adversarial methods for \IFO~in order to increase the data efficiency of these methods without sacrificing performance.
	Specifically, we consider time-varying linear Gaussian policies, and propose a method that integrates the linear-quadratic regulator with path integral policy improvement into an existing adversarial \IFO~framework.
	The result is a more data-efficient \IFO~algorithm with better performance, which we show empirically in four simulation domains: using far fewer interactions with the environment, the proposed method exhibits similar or better performance than the existing technique.
\end{abstract}

\section{Introduction}\label{sec:introduction}
Reinforcement learning (\RL) \cite{sutton1998reinforcement} is a machine learning paradigm that makes it possible for artificial, autonomous agents to learn tasks using their own experience.
Broadly speaking, \RL~agents repeatedly observe the state of the environment they inhabit, perform an action within that environment, and receive feedback based on the utility of taking that action with respect to a particular task that they are trying to perform.
Agents learn how to successfully perform tasks by modifying their action-selection policy so as to induce behavior that optimizes the expected cumulative feedback they receive.
While \RL~methods have led to artificial agents successfully learning to perform all kinds of tasks, the problem of designing good feedback, or cost, functions is still one that must be solved by humans, and doing so is difficult since it typically requires a great deal of domain knowledge.

\begin{figure}
	\centering
	\begin{tikzpicture}[scale=0.9, every node/.style={scale=0.9}]
	\draw [fill={rgb:blue,1;green,1;white,7}, rounded corners] (0,0) rectangle (5,5);
	\node [text=black, align=center] at (2.5,-.7) {\GAIfO};
	\draw [fill={rgb:blue,1;green,1;white,7}, rounded corners] (5.5,0) rectangle (8,7);
	\node [text=black, align=center] at (6.75,-.7) {\PILQR};
	\draw [fill={rgb:blue,1;green,0;white,9}, rounded corners] (1.3,1.5) circle (1.1);
	\node [text=black, align=center] at (1.3,1.5) {Expert\\Demonstration\\$D^e$};
	\draw [fill={rgb:blue,1;green,0;white,9}, rounded corners] (3.7,1.5) circle (1.1);
	\node [text=black, align=center] at (3.7,1.5) {Sampled\\Trajectories\\$\{\tau^i\}$};
	\draw [thick,-] (1.3,2.6) -- (1.3,3) -- (3.7,3) -- (3.7,2.6);
	\draw [thick,->] (2.5,3) -- (2.5,3.5);
	\draw [fill={rgb:blue,1;green,0;white,9}, rounded corners] (.5,3.5) rectangle (4.5,4.5);
	\node [text=black, align=center] at (2.5,4) {Update Discriminator\\$D_\theta = h\circ g_\theta$};
	\draw [thick,->] (2.5,4.5) -- (2.5,5.5);
	\draw [fill={rgb:blue,1;green,0;white,9}, rounded corners] (.5,5.5) rectangle (4.5,6.5);
	\node [text=black, align=center] at (2.5,6) {Extract Cost Function\\$c(s_t,a_t)$};
	\draw [thick,->] (4.5,6) -- (5.7,6);
	\draw [fill={rgb:blue,1;green,0;white,9}, rounded corners] (5.7,5.2) rectangle (7.8,6.8);
	\node [text=black, align=center] at (6.75,6) {Update\\Controller\\$p(a|s)$};
	\draw [thick,->] (6.75,5.2) -- (6.75,4.6);
	\draw [fill={rgb:blue,1;green,0;white,9}, rounded corners] (5.7,2.6) rectangle (7.8,4.6);
	\node [text=black, align=center] at (6.75,3.6) {Collect\\Trajectory\\Samples\\$\{\tau^i\}$};
	\draw [thick,->] (5.7,3.6) -- (5.25,3.6) -- (5.25,1.5) -- (4.8,1.5);
	\draw [thick,->] (6.75,2.6) -- (6.75,2);
	\draw [fill={rgb:blue,1;green,0;white,9}, rounded corners] (5.7,.2) rectangle (7.8,2);
	\node [text=black, align=center] at (6.75,1.1) {Fit\\Gaussian\\Dynamics\\$P(s'|s,a)$};
	\draw [thick,->] (7.8,1.1) -- (8.2,1.1) -- (8.2,6) -- (7.8,6);
	\draw [thick,->] (6.75,.2) -- (6.75,-.2) -- (-.2,-.2) -- (-.2,6) --(.5,6);
	\end{tikzpicture}
	\caption{A block diagram representing the learning flow of \DEALIO. A time-varying Gaussian controller $p(a|s)$ is initialized and used to collect trajectories $\{\tau^i\}$ which are then used to fit a linear Gaussian dynamics model $P(s'|s,a)$. The collected data $\{\tau^i\}$ is also used with the demonstration data $D^e$ to train a discriminator $D_\theta$ which is a composition function of a quadratic function $h(.,s_t, s_{t+1})$ and a neural network $g_\theta(s_t, s_{t+1})$. The discriminator $D_\theta$ and the dynamics model $P(s'|s,a)$ are then used to extract the cost function $c(s,a)$ which is then used to update the controller $p(a|s)$.}
	\label{fig:MDP}
\end{figure}
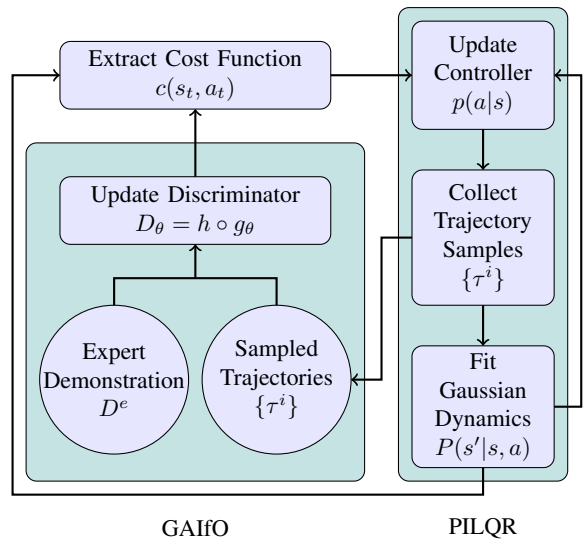

The paradigm of imitation learning (\IL) \cite{abbeel2004apprenticeship,Argall:2009:SRL:1523530.1524008,memarian2020active}, on the other hand, sidesteps the cost function design problem by instead using {\em demonstrations} of the desired behavior in order to guide the machine learning process.
Demonstrating tasks is often much easier for humans compared to designing a function that can provide detailed feedback to the agent in arbitrary situations.
In traditional \IL, however, the demonstration data must include information about both the states and actions experienced by the demonstrator. The requirement of actions in particular prevents the agent from being able to learn from cheaper, passive resources such as YouTube videos, which typically contain only state information in the form of {\em observations} of the demonstrator.

To remove the need for action information, the research community has recently given a great deal of attention to the related paradigm of imitation learning from observation (\IFO) \cite{DBLP:journals/corr/LiuGAL17,torabi2019recent}, which considers explicitly the case where artificial agents seek to learn behaviors from demonstrations consisting only of state information.
One class of algorithms that has achieved state-of-the-art performance on \IFO~problems relies on techniques from adversarial learning \cite{DBLP:journals/corr/MerelTTSLWWH17,torabi2019generative,sun2019provably}.
For example, generative adversarial imitation from observation (\GAIfO) \cite{torabi2019generative,torabi2019adversarial} uses adversarial learning to bring the state-transition distribution of the imitator closer to that of the demonstrator.
As with other adversarial \IFO~approaches, \GAIfO~relies on a model-free \RL~algorithm as part of the learning process, in which the goal is to learn a policy directly from samples without forming or leveraging any intermediate representations of the environment or cost function.
These model-free techniques typically exhibit extremely high sample complexity, which can prove problematic in real-world settings in which each sample required for learning incurs a high cost in terms of time, energy, and/or risk.

One classical way in which the \RL~community has dealt with poor sample efficiency is through the use of model-based \RL.
Unlike model-free techniques, model-based \RL~algorithms make assumptions about the environment dynamics such as it being linear or differentiable. and learn a model of the dynamics during the learning process.
Model-free \RL~algorithms are known to perform well at the cost of data inefficiency \cite{kober2013reinforcement,DBLP:journals/corr/SchulmanLMJA15}, i.e., the final learned policy is able to perform close to optimality, but learning it requires many interactions with the environment.
On the other hand, model-based algorithms are known to be sample-efficient, but, since they rely on assumptions to build models, they typically result in sub-optimal policies, especially in environments that exhibit complex dynamics \cite{deisenroth2013survey}.
Some \RL~algorithms, such as \PILQR~\cite{chebotar2017combining}, have been proposed to take advantage of the benefits of both model-based and model-free algorithms.

In this paper, we propose to address sample inefficiency in adversarial \IFO~algorithms by integrating ideas from model-based \RL.
In particular, we propose to integrate the sample-efficient \RL~updates from \PILQR~with the high-performing \GAIfO~algorithm for \IFO.
The resulting algorithm, Data-Efficient Adversarial Learning for Imitation from Observation (\DEALIO), is able to learn a time-varying linear Gaussian controller that can imitate a demonstrator using state-only demonstrations of a behavior.
We evaluate \DEALIO~in several simulation domains and compare it with \GAIfO.
Our results show that \DEALIO~can achieve good imitation performance using far fewer environment interactions during learning.
Moreover, in some cases, \DEALIO~is even able to outperform \GAIfO.\footnote{An earlier version of this work was presented in ICML Workshop on Imitation, Intent, and Interaction (I3) \cite{torabi2019sample}. This paper however, is drastically different both in the design of the proposed algorithm and the experiments.}

\section{Related Work}

We now review related work in the areas of sample-efficient reinforcement learning and imitation learning from observation.

\subsection{Reinforcement Learning}

Broadly speaking, reinforcement learning algorithms can be categorized as model-based or model-free. Model-free algorithms are known for their ability to handle complex dynamics \cite{kober2013reinforcement,DBLP:journals/corr/SchulmanLMJA15}, while model-based algorithm are known for their data efficiency \cite{deisenroth2013survey,tassa2012synthesis,chebotar2017combining}. With the goal of trying to reap the advantages of both techniques, many algorithms have been proposed that seek to combine the advantages of each \cite{farshidian2014learning,heess2015learning}. The particular work that \DEALIO~is most related to is \PILQR~which combines path integral policy improvement (\PI) \cite{theodorou2010generalized} and the linear-quadratic regulator (\LQR) \cite{tassa2012synthesis}.

\DEALIO~is similar to the work mentioned above in that it is intended to learn high quality policies with small number of interactions with the environment. However, it is different in that the methods mentioned above are reinforcement learning algorithms and require a cost function; while \DEALIO~is an \IFO~algorithm that imitates from a demonstrator.

\subsection{Imitation Learning from Observation (\IFO)}
\IFO~is a task faced by machine learners that seek to imitate state-only behavior demonstrations \cite{torabi2019recent}.
This topic has received a great deal of attention over the years, and several different types of \IFO~algorithms have been developed \cite{IJCAI2018-torabi,DBLP:journals/corr/LiuGAL17,DBLP:journals/corr/SermanetLHL17,pavse2020ridm}.
The work presented here is most related to adversarial imitation from observation algorithms.
These algorithms typically attempt to find imitation policies that induce behaviors that result distribution matching with the demonstrator with respect to either {\em states} \cite{DBLP:journals/corr/MerelTTSLWWH17,DBLP:journals/corr/abs-1709-06683} or {\em state transitions} \cite{torabi2019generative,torabi2019adversarial}.
One algorithm in particular that our work builds off of is Generative Adversarial Imitation from Observation (\GAIfO) \cite{torabi2019generative,torabi2019adversarial}, which exhibits good final performance on benchmark tasks, but also suffers from poor sample complexity.
Other \IFO~work includes that which explores the choice of imitator state representation \cite{torabi2019imitation} and work that has explored cases when the imitator and demonstrator exhibit different action spaces \cite{zolna2018reinforced} or viewpoints \cite{DBLP:journals/corr/StadieAS17}.

The algorithm we propose here, \DEALIO, is similar to the ones mentioned above in that our algorithm is also an adversarial imitation from observation algorithm.
However, \DEALIO~seeks to resolve high sample complexity issues by leveraging data-efficient reinforcement learning algorithms. In this respect, our work is similar to very recent work on off-policy \IFO~\cite{NEURIPS2020_92977ae4}, though we pursue a fundamentally-different approach by considering on-policy techniques.



\section{Background}
The proposed method is developed in the specific context of several techniques within the existing literature, which we review here in detail.

\subsection{Reinforcement Learning}
Reinforcement learning problems are posed in the context of a Markov Decision Process (\MDP), i.e., a tuple $\M = \{\S, \A, P, c\}$, where $\S$ and $\A$ are state and action spaces, respectively, $P$ is a transition probability distribution, and $c$ is the cost function.
At a given time instant, a decision-making agent operating within $M$ finds itself in state $s \in \S$ and takes an action $a \in \A$ based on a policy $\pi : \S \rightarrow \A$.
As a result, agent moves to a new state $s' \in \S$ with probability $P(s'|s,a)$, at which point the agent also receives feedback $c(s,a)$ based on the cost function $c:\S \times \A \rightarrow \mathbb{R}$.
The goal of reinforcement learning agents is to use their own experience to learn a policy that results in a behavior that incurs minimal expected cumulative cost.

\subsubsection{The Linear-Quadratic Regulator (\LQR)} \label{sec:lqr}
The Linear-Quadratic Regulator (\LQR) \cite{li2004iterative,sontag2013mathematical} is a sample-efficient, model-based reinforcement learning algorithm that makes several strict assumptions regarding the features of the system such as the environment transition distribution and the cost function.
Specifically, \LQR~assumes that the environment dynamics are linear, i.e.,
\begin{align} \label{eq:linear-dynamics}
s_{t+1} = f(s_t, a_t) =  F_t \begin{bmatrix} s_t \\ a_t \end{bmatrix} + f_t \; ,
\end{align}
where $F_t$ and $f_t$ are known matrices and vectors, respectively.
\LQR~also assumes the cost function is quadratic, i.e.,
\begin{align}\label{eq:quadratic-cost}
c(s_t, a_t) = \frac{1}{2} \begin{bmatrix} s_t \\ a_t\end{bmatrix}^T C_t  \begin{bmatrix} s_t \\ a_t \end{bmatrix} + \begin{bmatrix} s_t \\ a_t\end{bmatrix}^T c_t \; ,
\end{align}
where $C_t$ and $c_t$ are also known matrices and vectors, respectively.
Under these assumptions, \LQR~seeks an optimal controller (policy) of the form
\begin{align} \label{eq:linear-controller}
a_t = K_ts_t + k_t \; ,
\end{align}
where $K_t$ and $k_t$ are functions of the model parameters $F_t$, $C_t$, $f_t$, and $c_t$ that can be computed for each time step using a procedure known as backward recursion.
\LQR~can also be used under the assumption of linear Gaussian dynamics, i.e., 
\begin{align} \label{eq:gaussian-dynamics}
s_{t+1} \sim P(s'|s,a) &= \N(f(s_t, a_t),\Sigma_t) \\ &=\N(F_t \begin{bmatrix} s_t \\ a_t \end{bmatrix} + f_t, \Sigma_t) \; .\notag
\end{align}
In this case, the solution is still a controller of the form specified in Equation \ref{eq:linear-controller}, where $K_t$ and $k_t$ can again be calculated using \LQR~backward recursion.

Note that, to compute the controllers as described above, one must know the dynamics model parameters $F_t$ and $f_t$.
If the dynamics model is unknown, methods have been proposed in which it can be learned through experience \cite{levine2014learning}.
However, for complex dynamics, one global model is generally not sufficient for the entire state space, and so local dynamics models are typically learned instead, i.e., new linear Gaussian dynamics models are fit to the available data for each time step.


\subsubsection{Combining Path Integral Policy Improvement and \LQR~(\PILQR)}\label{sec:pilqr}
In trying to derive the benefits of both model-free and model-based \RL~techniques, the \PILQR~algorithm \cite{chebotar2017combining} integrates model-based updates from \LQR~with fitted linear models and Path Integral Policy Improvement (\PI) \cite{theodorou2010generalized}, which is a model-free \RL~algorithm based on stochastic optimal control.
\PILQR~removes the quadratic cost requirement of \LQR, instead requiring only that the cost function be twice differentiable.
\PILQR~forms a quadratic approximation of the cost, and makes policy updates by first using the iterative Linear-Quadratic Regulator (\iLQR) approach \cite{li2004iterative} with the quadratic approximation, and then performing a subsequent policy update using \PI~on the residual cost.



\subsection{Imitation Learning}
Imitation learning (\IL) is a machine learning paradigm in which autonomous agents seek to learn behaviors without having access to explicit cost feedback as in \RL, instead operating in the context of a \MDP~without cost, $\M\backslash c$.
In this setting, the agent instead has access to some demonstrations of desirable behavior.
These demonstrations consist of the states and actions of the demonstrator, $D^e = \{\tau_i^e\}$, where $\tau^e_i = \{(s^e,a^e)\}_i$.
Requiring access to the actions of the demonstrator however, makes it impossible for the agents to learn from the cases where videos of the demonstrator are the only available resources.

\subsection{Imitation Learning from Observation (\IFO)} \label{sec:ifo}
Imitation learning from observation (\IFO) removes the requirement in \IL~that the demonstrations have to contain action information.
That is, in \IFO, demonstrations consist of state information only, i.e., $D^e = \{\tau_i^e\}$, where $\tau_i^e =\{(s^e)\}_i$.

Our work builds off of an \IFO~algorithm called Generative Adversarial Imitation from Observation (\GAIfO) \cite{torabi2019generative,torabi2019adversarial,torabi2019imitation} which is inspired by Generative Adversarial Networks (\GAN s) \cite{goodfellow2014generative}.
\GAIfO~attempts to learn a policy such that the imitator's state-transition distribution matches that of the demonstrator.
The algorithm works as follows.
First, a random neural network policy $\pi_\phi$ is initialized as the imitator's policy, which is used by the imitator to generate state trajectories $\{\tau_i^i\}$ where $\tau_i^i = \{(s^i)\}_i$.
Next, a neural network discriminator is trained to discriminate between the state transitions provided by the demonstrator and the state transitions generated by the imitator.
Specifically, the discriminator training is done using supervised learning with the following loss function:
\begin{align}\label{eq:disc-update}
 &-\Big(\mathbb{E}_{\{\tau^i_i\}} [\log(D_\theta(s_t,s_{t+1}))]\\&+\mathbb{E}_{\{\tau^e_i\}} [\log(1-D_\theta(s_t,s_{t+1}))]\Big) \; ,  \notag
\end{align}
where $D$ is the discriminator network parameterized by $\theta$ and $s_t$ and $s_{t+1}$ are two consecutive states.
Finally, the model-free \RL~algorithm Trust Region Policy Optimization (\TRPO) \cite{DBLP:journals/corr/SchulmanLMJA15} is used to update the imitator's policy using the cost function
\begin{align}\label{eq:gaifo-cost}
 \mathbb{E}_{\{\tau^i_i\}} [\log(D_\theta(s_t,s_{t+1}))] \; .
\end{align}
While \GAIfO~has shown very promising performance, it also exhibits high sample complexity.

\section{\DEALIO}

We now introduce \DEALIO, an algorithm for \IFO~that has the explicit goal of making adversarial imitation learning from observation techniques more sample efficient through the use of model-based \RL.
\DEALIO~is based on the \GAIfO~algorithm discussed in Section \ref{sec:ifo}, in which a neural network discriminator attempts to distinguish between state transitions generated by the current imitation policy and state transitions generated by the demonstrator.
The discriminator's output is used as the cost function that drives imitation policy learning using the model-free \RL~algorithm \TRPO.
To improve sample efficiency, \DEALIO~aims to replace \TRPO~with the \PILQR~algorithm that computes policy updates using a combination of model-based and model-free \RL.

Making this change to \GAIfO~is nontrivial due to the functional form of the cost function required by \PILQR.
First, \PILQR~requires a quadratic approximation of the cost function in order to compute updates, whereas the cost function used in \GAIfO~is specified by the discriminator, which is a continually updating deep neural network. As mentioned in Section \ref{sec:pilqr}, it is not required for the cost function to be in the form of Equation \ref{eq:quadratic-cost}, instead it only needs to be twice-differentiable.
Therefore, we aim to develop a cost function in the form of
\begin{align}\label{eq:pilqr-cost}
c(s_t, a_t) = \frac{1}{2} \begin{bmatrix} s_t \\ a_t\end{bmatrix}^T C_t  \begin{bmatrix} s_t \\ a_t \end{bmatrix} + \begin{bmatrix} s_t \\ a_t\end{bmatrix}^T c_t + cc_t \; .
\end{align}
for which the quadratic approximation would be the first two terms of the right hand side.\footnote{Another option is to use a neural network discriminator (the same as \GAIfO) and take the second order Taylor expansion of the neural network to calculate the quadratic approximate of the cost function. This approach also meets the requirements of \PILQR. We implemented this approach and saw that \DEALIO~outperforms it by a large margin. We posit that the reason is that the manifold learned using the neural network is very complex that its second order Taylor expansion is not meaningful enough.}
Second, \PILQR~requires a cost function over both states and actions (as suggested by Equation \ref{eq:pilqr-cost}), whereas, because \GAIfO~is an \IFO~technique, the discriminator that specifies the cost function is a function of state information only.

Overcoming these challenges and learning a cost function presented in Equation \ref{eq:pilqr-cost} requires us to find a method by which we can compute $C_t$, $c_t$, and $cc_t$ in the context of the \GAIfO~algorithmic structure.
To do so, we first propose to modify the structure of the discriminator by considering it to be a composition of two functions, $g_\theta(s_t, s_{t+1})$ and $h(.,s_t, s_{t+1})$.
The first function, $g_\theta(s_t, s_{t+1})$, is a neural network (with parameters $\theta$) that takes state transitions as inputs and outputs the following quantities:
\begin{itemize}
	\item The elements of a matrix $C^{ss}(s_t, s_{t+1})$
	\item The elements of a vector $c^{ss}(s_t, s_{t+1})$
	\item A constant $cc^{ss}(s_t, s_{t+1})$
\end{itemize}
The second function, $h(.,s_t, s_{t+1})$, is a quadratic function in which the outputs of $g_\theta(s_t,s_{t+1})$, i.e. $C^{ss}(s_t, s_{t+1})$ and $c^{ss}(s_t, s_{t+1})$, can be used as its matrix and vector and the overall function composition becomes
\begin{align}\label{eq:c-ss}
D_\theta(s_t,s_{t+1}) &= h(g_\theta(s_t, s_{t+1}), s_t, s_{t+1}) \\&= \frac{1}{2} \begin{bmatrix} s_t \\ s_{t+1}\end{bmatrix}^T C^{ss} (s_t, s_{t+1}) \begin{bmatrix} s_t \\ s_{t+1} \end{bmatrix} \notag\\
&+ \begin{bmatrix} s_t \\ s_{t+1}\end{bmatrix}^T c^{ss} (s_t, s_{t+1}) \; . \notag
\end{align}
which yield the discriminator we use in \DEALIO.
By imposing this structure on the discriminator, we now have a quadratic cost function, $h(g_\theta(s_t, s_{t+1}), s_t, s_{t+1})$, as required by \PILQR.
However, two issues still remain.
First, $h(g_\theta(s_t, s_{t+1}), s_t, s_{t+1})$ is still a function of state transitions rather than state-action pairs.
Second, $C^{ss}(s_t, s_{t+1})$ and $c^{ss}(s_t, s_{t+1})$ are functions of states rather than time.

In order to find a quadratic cost function with state-action pairs as input, we use the linear dynamics model assumed by \PILQR~in Equation \ref{eq:linear-dynamics}, which allows us to rewrite $s_{t+1}$ in $h(g_\theta(s_t, s_{t+1}), s_t, s_{t+1})$ as a function of $s_t$ and $a_t$.
To do so, we first partition quantities from Equations \ref{eq:linear-dynamics} and \ref{eq:c-ss} as follows:
\begin{align}
F_t = \begin{bmatrix} F_{s_t} \\ F_{a_t} \end{bmatrix} \; ,
\end{align}

\begin{align}
C^{ss} (s_t, s_{t+1}) = \begin{bmatrix} C^{ss}_{s_t,s_t} ~ ~ C^{ss}_{s_t,s_{t+1}} \\ C^{ss}_{s_{t+1},s_t} ~ ~ C^{ss}_{s_{t+1},s_{t+1}} \end{bmatrix} \; ,
\end{align}
and
\begin{align}
c^{ss} (s_t, s_{t+1}) = \begin{bmatrix} c^{ss}_{s_t} \\ c^{ss}_{s_{t+1}} \end{bmatrix} \; .
\end{align}
Next, substituting $s_{t+1}$ from Equation \ref{eq:linear-dynamics} into Equation \ref{eq:c-ss} and doing some linear algebra, we can write
\begin{align}
h(s_t, a_t) &= \frac{1}{2} \begin{bmatrix} s_t \\ a_t\end{bmatrix}^T C^{sa}(s_t, s_{t+1})  \begin{bmatrix} s_t \\ a_t \end{bmatrix} \\
&+ \begin{bmatrix} s_t \\ a_t\end{bmatrix}^T c^{sa}(s_t, s_{t+1})  \; , \notag
\end{align}
where the new matrix and vector, $C^{sa}(s_t, s_{t+1}) $ and $c^{sa}(s_t, s_{t+1}) $, respectively, are given by
\begin{align}
C^{sa}(s_t, s_{t+1})  = \begin{bmatrix} C^{sa}_{s_t,a_t} ~ ~ C^{sa}_{s_t,a_t} \\ C^{sa}_{a_t,s_t} ~ ~ C^{sa}_{a_t,a_t} \end{bmatrix} \; ,
\end{align}
and
\begin{align}
c^{sa}(s_t, s_{t+1})  = \begin{bmatrix} c^{sa}_{s_t} \\ c^{sa}_{a_t} \end{bmatrix} \; ,
\end{align}
and the individual partition terms above are computed as
\begin{align}
C^{sa}_{s_t,s_t} &= C^{ss}_{s_t,s_t} + F_{s_t}^T C^{ss}_{s_{t+1},s_t} + C^{ss}_{s_t,s_{t+1}} F_{s_t}  \\&+ F_{s_t}^T C^{ss}_{s_{t+1},s_{t+1}} F_{s_t} \; ,\notag
\end{align}

\begin{align}
C^{sa}_{s_t,a_t} = C^{ss}_{s_t,s_{t+1}} F_{a_t} + F_{s_t}^T C^{ss}_{s_{t+1},s_{t+1}} F_{a_t} \; ,
\end{align}

\begin{align}
C^{sa}_{a_t,s_t} = F_{a_t}^T C^{ss}_{s_{t+1},s_t} + F_{a_t}^T C^{ss}_{s_{t+1},s_{t+1}} F_{s_t} \; ,
\end{align}

\begin{align}
C^{sa}_{a_t,a_t} =  F_{a_t}^T C^{ss}_{s_{t+1},s_{t+1}} F_{a_t} \; ,
\end{align}

\begin{align}
c^{sa}_{s_t} &= \frac{1}{2} C^{ssT}_{s_{t+1},s_t} f_t +\frac{1}{2} C^{ss}_{s_t,s_{t+1}} f_t + \frac{1}{2} F_{s_t}^T C^{ssT}_{s_{t+1},s_{t+1}} f_t \\
&+ \frac{1}{2} F_{s_t}^T C^{ss}_{s_{t+1},s_{t+1}} f_t + c^{ss}_{s_t} + F_{s_t}^T c^{ss}_{s_{t+1}} \; ,\notag
\end{align}
and
\begin{align}
c^{sa}_{a_t}  = \frac{1}{2} F_{a_t}^T C^{ssT}_{s_{t+1},s_{t+1}} f_t + \frac{1}{2} F_{a_t}^T C^{ss}_{s_{t+1},s_{t+1}} f_t + F_{a_t}^T c^{ss}_{s_{t+1}} \; .
\end{align}

Finally, in order to find cost function parameters $C_t$, $c_t$, and $cc_t$ (as shown in Equation \ref{eq:pilqr-cost}) to be functions of time only (i.e., independent of states), we use the mean state transition at time $t$ as input to $C^{sa}(\cdot,\cdot)$, $c^{sa}(\cdot, \cdot)$, and $cc^{ss}(\cdot, \cdot)$, i.e.,
\begin{align}
C_t = C^{sa} (\bar{s}_t, \bar{s}_{t+1}) \; ,
\end{align}
\begin{align}
c_t = c^{sa} (\bar{s}_t, \bar{s}_{t+1}) \; ,
\end{align}
and
\begin{align}
cc_t = cc^{ss} (\bar{s}_t, \bar{s}_{t+1}) \; ,
\end{align}
where $\bar{s}_t$ represents the mean taken over all the available sample states at each time step.\footnote{One alternative is to calculate the average of $C^{sa}(s_t, s_{t+1})$ and $c^{sa}(s_t, s_{t+1})$ over the available states, however, our experiments showed that calculating $C^{sa} (\bar{s}_t, \bar{s}_{t+1})$ and $c^{sa} (\bar{s}_t, \bar{s}_{t+1})$ results in better performance.}
With the above quantities defined, the final cost function (as presented in Equation \ref{eq:pilqr-cost}) used to perform \PILQR~updates in \DEALIO~is prepared.

\begin{algorithm}[t!]
	\caption{\DEALIO}\label{alg:gaifo-pilqr}
	\begin{algorithmic}[1]
		\State Initialize controller $p(a|s)$\label{alg:in-p}
		\State Initialize a neural network discriminator $D_\theta$ with random parameter $\theta$\label{alg:in-d}
		\State Collect demonstration trajectories $D^e = \{\tau^e\}$
		\While {Controller Improves}
		\State Execute the controller to collect state-action trajectories $\{\tau^i_i\}$\label{alg:ex-controller}
		\State Fit a linear Gaussian dynamics model $P(s'|s,a)$\label{alg:fit-dyn}
		\State Update $D_\theta$ using loss \begin{align}&-\Big(\mathbb{E}_{\{\tau^i_i\}} [\log(D_\theta(s_t,s_{t+1}))]\notag\\&+\mathbb{E}_{\{\tau^e_i\}} [\log(1-D_\theta(s_t,s_{t+1}))]\Big)\notag\end{align}
		and store $C^{ss} (s_t, s_{t+1})$, $c^{ss} (s_t, s_{t+1})$, and $cc^{ss}(s_t, s_{t+1})$\label{alg:tr-disc}
		\State Compute the cost $c(s_t, a_t)$ by calculating $C_t$, $c_t$, and $cc_t$ \label{alg:cost}
		\State Perform a \PILQR~update to update the controller $p(a|s)$ \label{alg:pilqr}
		\EndWhile
	\end{algorithmic}
\end{algorithm}

\begin{figure*}[t!]
	\centering
	\subfigure[Disc]{\label{fig:disc}\includegraphics[scale=0.09]{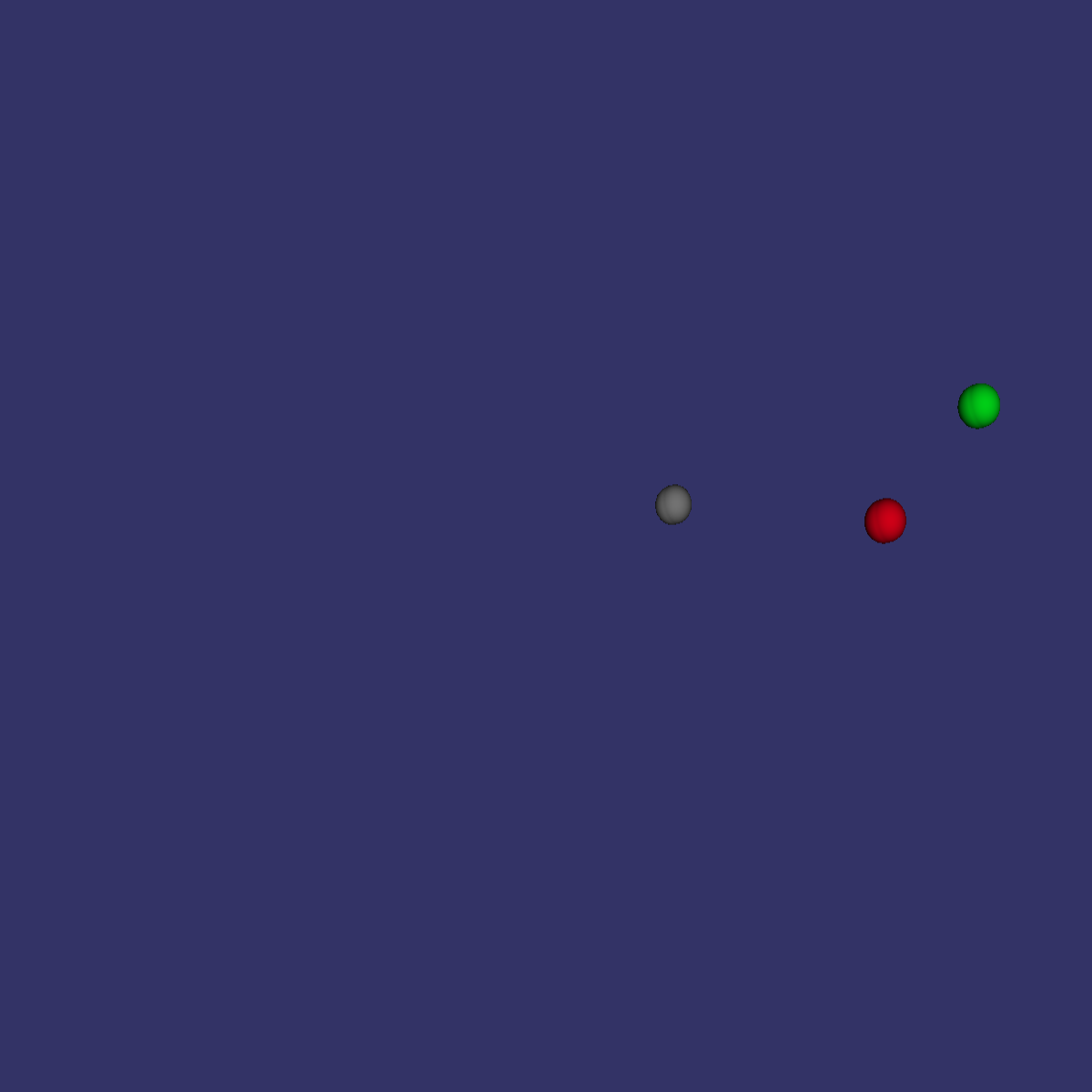}}
	\subfigure[PegInsertion]{\label{fig:peg}\includegraphics[scale=0.09]{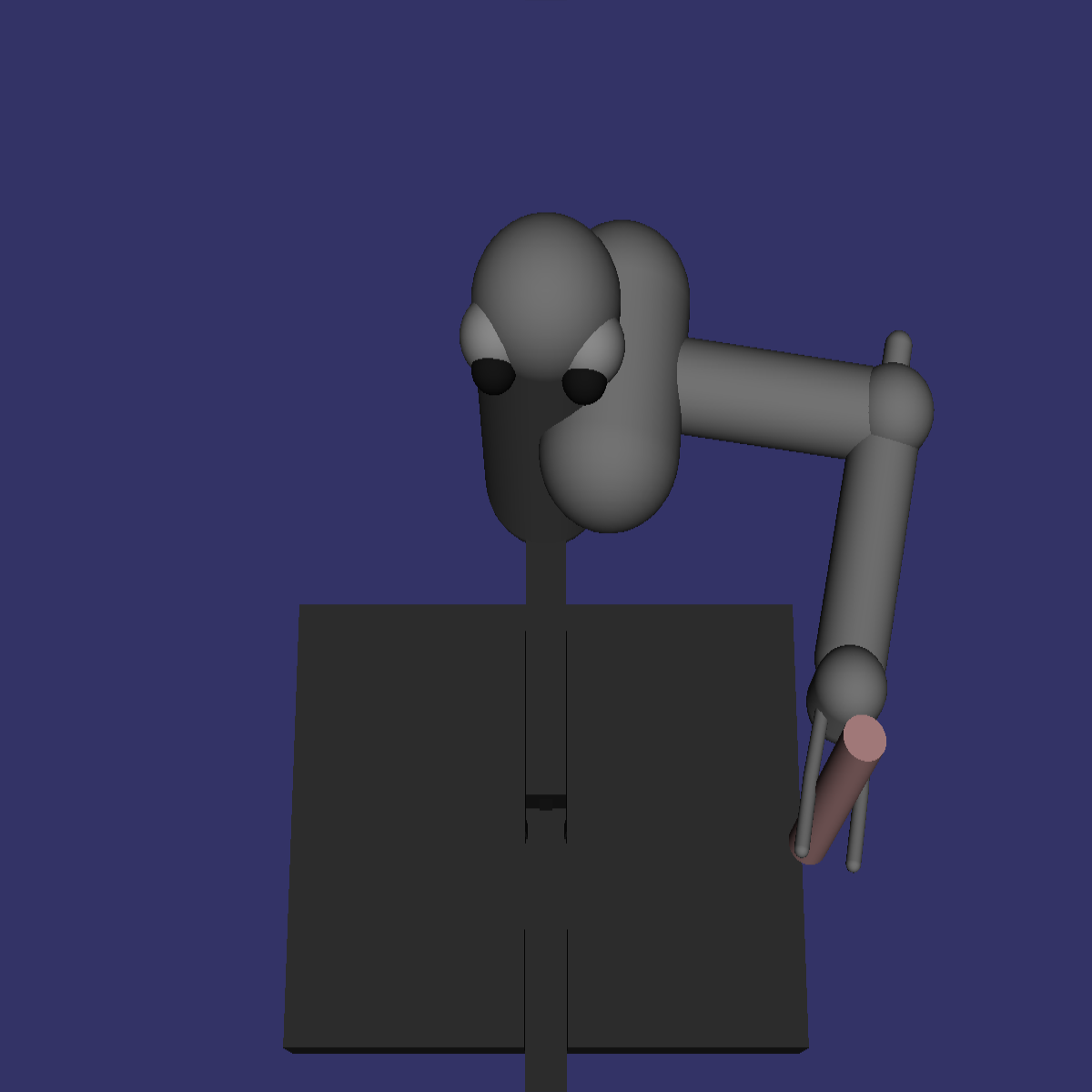}}
	\subfigure[GripperPusher]{\label{fig:gripper}\includegraphics[scale=0.09]{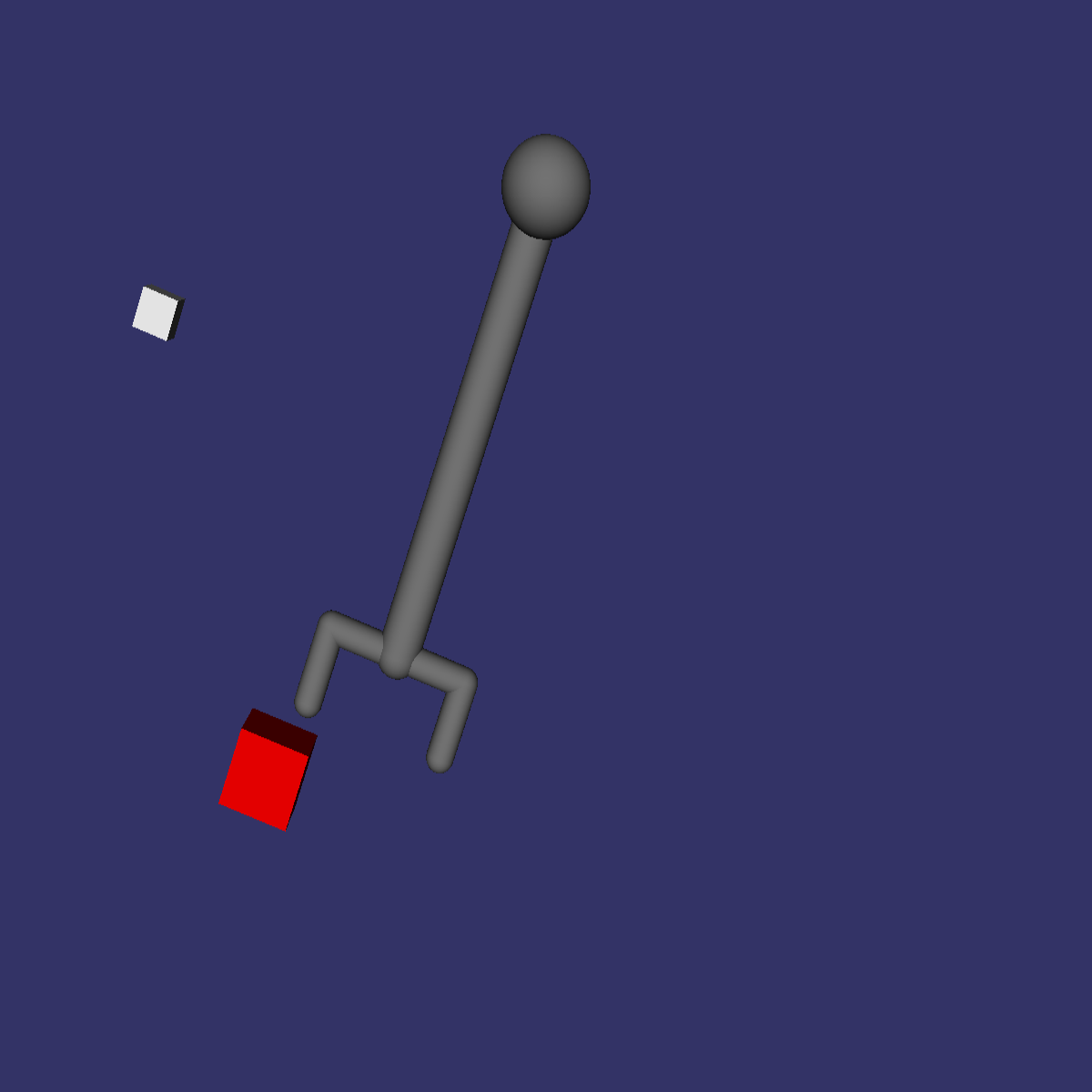}}
	\subfigure[DoorOpening]{\label{fig:door}\includegraphics[scale=0.09]{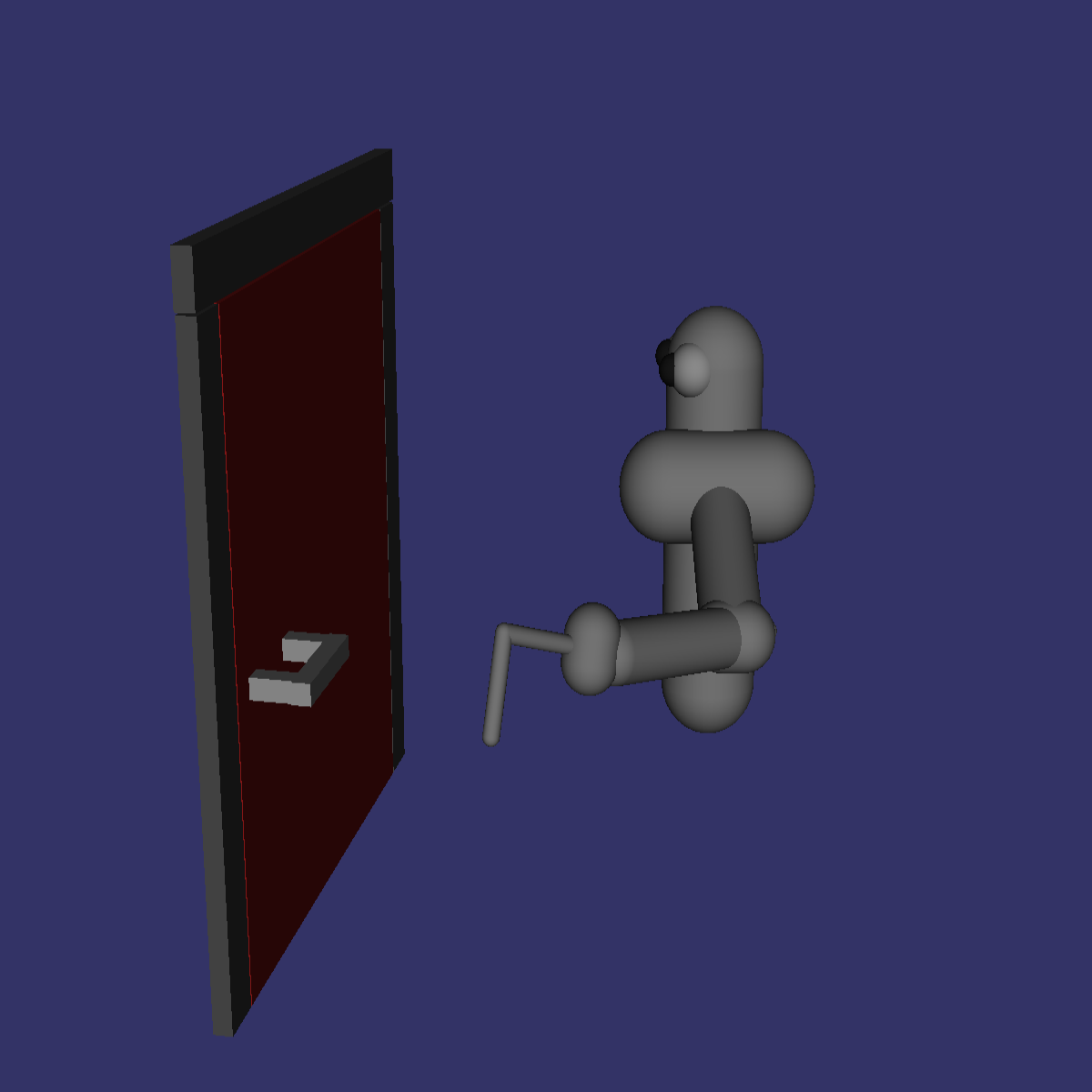}}
	\caption{Representative screenshots of the MuJoCo domains considered in this paper.}
	\label{fig:mujoco}
\end{figure*}

\begin{figure*}[ht]
	\centering
	\includegraphics[scale=.5]{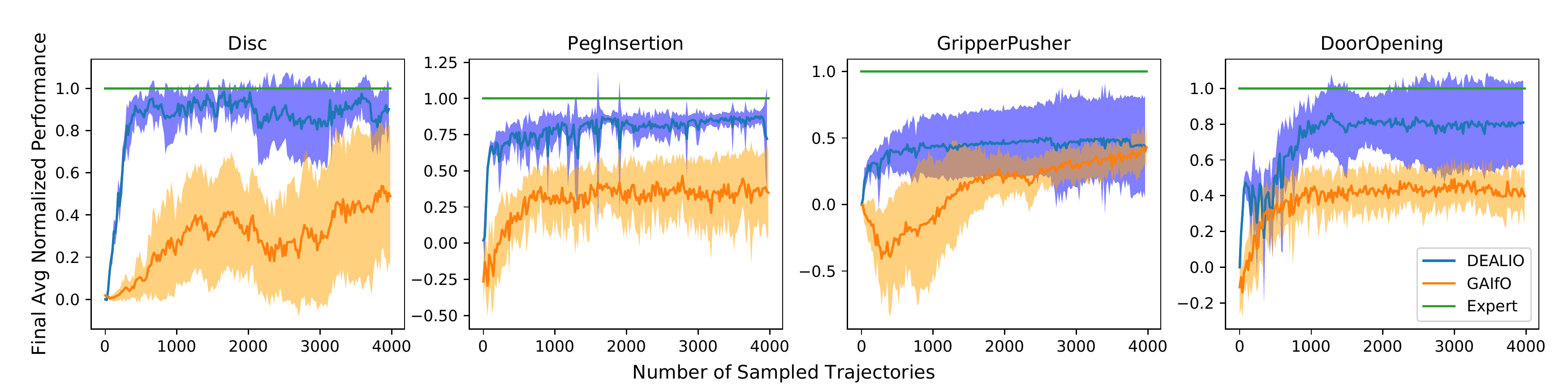}
	\caption{Learning with \DEALIO~is compared against \GAIfO. The plots represent the performance of the algorithms with respect to the number of trajectories sampled during the learning process. The solid lines represent the mean of the performances over 10 different training processes and the shaded areas represent the standard deviations. For comparison purposes, all the performances are scaled such that a random and the expert policy score $0.0$ and $1.0$, respectively.}
	\label{fig:dealio performance}
\end{figure*}

\vspace{0.2in} \noindent \textbf{\DEALIO} Having resolved the incompatibility between \GAIfO~and \PILQR~above, we now describe the full proposed algorithm, \DEALIO, detailed in Algorithm \ref{alg:gaifo-pilqr}.
First, we initialize a time-varying, linear Gaussian imitation controller $p(a|s)$ and a neural network discriminator $D_\theta$ (Lines \ref{alg:in-p} and \ref{alg:in-d}).
The imitator uses $p$ to generate multiple state-action trajectories $\tau^i = \{(s^i, a^i)\}$ (Line \ref{alg:ex-controller}), and these trajectories are used to fit a linear Gaussian dynamics model $P(s'|s,a)$, as shown in Equation \ref{eq:gaussian-dynamics} (Line \ref{alg:fit-dyn}).
Next, state transitions from both the imitator and the demonstrator are used to train the discriminator $D_\theta$ (Equation \ref{eq:c-ss}) using the same loss function as in \GAIfO~(Line \ref{alg:tr-disc}).
Then, $h$ and $P$ are used to calculate the quadratic cost function parameters in Equation \ref{eq:pilqr-cost} (Line \ref{alg:cost}).
Finally, the cost function is used to perform a \PILQR~update (Line \ref{alg:pilqr}).

\section{Experiments}
In order to evaluate \DEALIO, we ran several experiments using a benchmark robotics simulation environment.
In particular, we sought to determine whether or not \DEALIO, which integrates a model-based \RL~technique with the \GAIfO~algorithm for \IFO, exhibited better data efficiency than the original \GAIfO~algorithm, which instead uses a model-free \RL~technique.
We compared the data efficiency of each approach by analyzing learning curves---more data-efficient approaches exhibit steeper learning curves (i.e., they rise to a higher level of task performance using fewer samples).
As we will detail below, our results do indeed confirm that \DEALIO~is more data-efficient than \GAIfO, suggesting that integrating model-based \RL~approaches is a promising path forward for designing \IFO~approaches that are practical for real-world systems.
Moreover, we also found that, even given fewer training samples, \DEALIO~also led to better overall task performance than \GAIfO~in our experimental tasks.

\subsection{Experimental Setup}
We conducted our experiments using four robotics tasks simulated using the MuJoCo physics engine \cite{DBLP:conf/iros/TodorovET12}:
\begin{itemize}
	\item \textbf{Disc}: This domain includes a gray point particle on a disc and two target points---one red and one green---as shown in Fig. \ref{fig:disc}. The agent can push the particle in the x and y directions, and the task is for the agent to first move the particle to the red target point, and then move the particle to the green target point.  The state and action spaces are ten- and two-dimensional, respectively.
	\item \textbf{PegInsertion}: This domain includes an arm with a gripper, a peg, and a plate with a hole in it as shown in Fig. \ref{fig:peg}. The task is for the agent to manipulate the arm such that the peg is inserted into the hole. The state and action spaces are twenty-six- and seven-dimensional, respectively. 
	\item \textbf{GripperPusher}: This domain includes an arm with a gripper, a white particle, and a red target point as shown in Fig. \ref{fig:gripper}. The task is for the agent to manipulate the arm such that it reaches the particle, grabs it, and moves it to the target point. The state and action spaces are thirty-six- and four-dimensional, respectively.
	\item \textbf{DoorOpening}: This domain includes a robot arm and a door with a handle as shown in in Fig. \ref{fig:door}. The task is for the agent to manipulate the arm such that it reaches the handle and opens the door. The state and action spaces are thirty-six- and six-dimensional, respectively. 
\end{itemize}

In order to generate demonstration data $D^e$, expert demonstrators are trained for each task until convergence using \PILQR~with predefined task cost functions that have been previously used in related work \cite{leonetti2012combining,montgomery2016guided}. The trained policies are then used to generate 20 demonstration trajectories for each task.
The performance of these expert agents is represented as the green horizontal lines in Fig. \ref{fig:dealio performance}, where the performance metric (expected cumulative reward) is normalized to the expert's level.


\subsection{Implementation Details}
In our experiments, the neural network component of the \DEALIO~discriminator was modeled with two hidden layers, each with one hundred neurons. For fair comparison, the \GAIfO~discriminator was similarly modeled.
One important hyperparameter that must be specified for both \DEALIO~and \GAIfO~is the number of trajectory samples generated by the imitator at each iteration (e.g., for \DEALIO, Line \ref{alg:ex-controller} of Algorithm \ref{alg:gaifo-pilqr});  we empirically determined the best value of this hyperparameter separately for each algorithm and environment.
For \DEALIO, we collect ten, three, twenty, and twenty trajectories (each trajectory is one hundred time steps) for each domain as ordered in Fig. \ref{fig:mujoco}.
For \GAIfO, these same parameters were set as five, five, twenty, and twenty with the same order again.
Another important set of hyperparameters required by both algorithms is the number of updates to make to both the discriminator and the controller at each iteration.
Again, we empirically determined the best values for these hyperparameters, and we found that ten discriminator updates per iteration and one update for the controller per iteration worked best regardless of algorithm or domain.
Above, ten discriminator updates means that we divide the overall data collected at that iteration into ten batches and updated the discriminator once for each batch.

\subsection{Results and Discussion}
For each environment and algorithm, we generated the learning curves shown in Fig. \ref{fig:dealio performance} by computing the mean and standard deviation of the developing imitation policies over ten independent training runs.
The results confirm our hypothesis that integrating a model-based \RL~algorithm leads to improved data efficiency: the \DEALIO~learning curves all exhibit a significantly steeper rise than those obtained with \GAIfO.
The results also show that, even over much longer training horizons, \DEALIO~still achieves better final performance than \GAIfO.
We posit that this is possible because \PILQR~itself combines model-free and model-based reinforcement learning algorithms such that it is both data efficient and high performing.
Taken together, our results suggest that model-based \RL~can play an important role in bringing adversarial \IFO~methods to real-world scenarios.

\section{Conclusion and Future Work}
In this paper, we studied the effect of integrating sample-efficient, model-based \RL~techniques with recent adversarial \IFO~algorithms.
To do so, we introduced one way to perform this integration through our novel algorithm, \DEALIO, and demonstrated experimentally that (1) it exhibits significantly reduced sample complexity compared to \GAIfO, and (2) it can do so seemingly without sacrificing performance.

While our work suggests that model-based \RL~can play an important role in improving adversarial techniques for \IFO, there are several ways in which we might improve the exemplar we proposed here.
One in particular concerns our use of \PILQR, which is a trajectory-centric reinforcement learning algorithm, i.e., it finds a local controller, $p(a|s)$, only for a very specific task with a very specific initial state and goal.
While this approach appears to have been sufficient for our experimental domains, an interesting direction for future work is to take advantage of algorithms such as Guided Policy Search (\mbox{GPS}) \cite{levine2014learning} to integrate general neural network policy learning using a supervised learning algorithm to mimic the combined behavior of all the learned local controllers for different initial states and goals.

%
%
%
%
%
%
%
%

\bibliographystyle{IEEEtran}
\bibliography{IEEEabrv,icra2019}

\addtolength{\textheight}{-12cm}   

\end{document}